\documentclass[10pt,twocolumn,letterpaper]{article}

\usepackage{cvpr}
\usepackage{times}
\usepackage{epsfig}
\usepackage{graphicx}
\usepackage{amsmath}
\usepackage{amssymb}
\usepackage{multirow}

\begin{document}

\title{SYN-MAD 2022: Competition on Face Morphing Attack Detection Based on Privacy-aware Synthetic Training Data}

\author{\parbox{16cm}{\centering
    {\large Marco Huber$^{1}{}^{,}{}^{2}{}^{,}{}^{*}{}$, 
    Fadi Boutros$^{1}{}^{,}{}^{2}{}^{,}{}^{*}{}$, 
    Anh Thi Luu$^{1}{}^{,}{}^{*}{}$,
    Kiran Raja$^{3}{}^{,}{}^{*}{}$, 
    Raghavendra Ramachandra$^{3}{}^{,}{}^{*}{}$, 
    Naser Damer$^{1}{}^{,}{}^{2}{}^{,}{}^{*}{}$, \\ \vspace{1mm}
    Pedro C. Neto$^{4}{}^{,}{}^{5}{}^{,}{}^{+}{}$,
    Tiago Gon\c{c}alves$^{4}{}^{,}{}^{5}{}^{,}{}^{+}{}$,
    Ana F. Sequeira$^{4}{}^{,}{}^{5}{}^{,}{}^{+}{}$,
    Jaime S. Cardoso$^{4}{}^{,}{}^{5}{}^{,}{}^{+}{}$,
    Jo\~{a}o Tremo\c{c}o$^{6}{}^{,}{}^{+}{}$,
    Miguel Louren\c{c}o$^{6}{}^{,}{}^{+}{}$,
    Sergio Serra$^{7}{}^{,}{}^{+}{}$,
    Eduardo Cerme\~{n}o$^{7}{}^{,}{}^{+}{}$,
    Marija Ivanovska$^{8}{}^{,}{}^{+}{}$,
    Borut Batagelj$^{8}{}^{,}{}^{+}{}$,
    Andrej Kronov\v{s}ek$^{8}{}^{,}{}^{+}{}$,
    Peter Peer$^{8}{}^{,}{}^{+}{}$,
    Vitomir \v{S}truc$^{8}{}^{,}{}^{+}{}$ 
    }\\
    {\normalsize
    $^{1}$Fraunhofer Institute for Computer Graphics Research IGD, Germany - 
    $^{2}$TU Darmstadt, Germany - 
    $^{3}$Norwegian University of Science and Technology, Norway -
    $^{4}$INESC TEC, Portugal  -
    $^{5}$University of Porto, Portugal 
    $^{6}$Yoonik, Portugal - 
    $^{7}$Vaelsys R\&D, Spain -
    $^{8}$University of Ljubljana, Slovenia\\
    $^{*}$Competition organizer
    $^{+}$Competition participant\\
    Email: marco.huber@igd.fraunhofer.de}}
    }

\maketitle
\thispagestyle{empty}

\begin{abstract}
This paper presents a summary of the Competition on Face Morphing Attack Detection Based on Privacy-aware Synthetic Training Data (SYN-MAD) held at the 2022 International Joint Conference on Biometrics (IJCB 2022). The competition attracted a total of 12 participating teams, both from academia and industry and present in 11 different countries. In the end, seven valid submissions were submitted by the participating teams and evaluated by the organizers. The competition was held to present and attract solutions that deal with detecting face morphing attacks while protecting people's privacy for ethical and legal reasons. To ensure this, the training data was limited to synthetic data provided by the organizers. The submitted solutions presented innovations that led to outperforming the considered baseline in many experimental settings. The evaluation benchmark is now available at: https://github.com/marcohuber/SYN-MAD-2022.

\end{abstract}
\vspace{-5mm}
\section{Introduction}
\vspace{-2mm}
Current face recognition (FR) systems are vulnerable to different kinds of attacks, including so-called morphing attacks \cite{DBLP:conf/icb/FerraraFM14}. Morphing attacks combine two face images of two different individuals so that the resulting image can be, automatically or by human operators, matched to both individuals. If these morphed images are used for identity or travel documents, they would enable multiple subjects to be verified by the same documents, resulting in enormous security risks. To reduce this risk and counter these kinds of attacks, morphing attack detectors (MAD) \cite{NIST_Morph} are developed to detect morphed images beside natural, unaltered bona fide images. These morphing attack detectors are often trained on morphed images based on real person images, raising the question of whether the privacy of the individual is respected as the given consent is often questioned \cite{DBLP:journals/corr/abs-2203-06691}. Furthermore, the amount of morphs for training is often limited in terms of quantity and quality. There was previously a series of competitions on face presentation attacks that focused on spoofing attacks \cite{DBLP:conf/icb/ChingovskaYLYLKGDKNKPGKBRKGWZAKGTPSRGFPPSRAM13,DBLP:conf/icb/PurnapatraSBDYM21}. However, this is the first competition on morphing attack detection and the first competition targeting any attack on face recognition system while limiting its development data to synthetic data.

The need for solutions to detect morphing attacks arose after uncovering the threat of morphing attacks \cite{DBLP:conf/icb/FerraraFM14} and the vulnerability of face recognition systems to these types of attacks \cite{DBLP:conf/biosig/ScherhagNRGVSSM17,DBLP:conf/btas/DamerS0K18,DBLP:conf/icb/DamerSZWTKK19}. As a result, a number of datasets for developing MAD systems were created \cite{DBLP:conf/btas/RaghavendraRB16, DBLP:conf/cvpr/RaghavendraRVB17a, DBLP:conf/icb/RaghavendraRVB17,DBLP:conf/isvc/DamerSFBKK21,DBLP:conf/isvc/DamerRSVBFKRK21,DBLP:conf/dagm/DamerBWBTBK18}, which all used the privacy/legal/ethical sensitive face images of real individuals and were limited in quantity and variation.
Most of these dataset consists of morph images based on interpolation of facial landmarks \cite{DBLP:conf/btas/RaghavendraRB16, DBLP:journals/tifs/ScherhagRMB20,DBLP:journals/tifs/FerraraFM18,DBLP:conf/isvc/DamerSFBKK21}. More recently the use of datasets based on generative adversial networks (GAN) gained attention \cite{DBLP:conf/btas/DamerS0K18,DBLP:conf/btas/DamerBSKK19,DBLP:conf/iciap/DebiasiDSRSBKU19,DBLP:conf/btas/DamerGZKK19,DBLP:journals/tbbis/ZhangVRRDB21, DBLP:conf/iwbf/VenkateshZRRDB20}.

Most of the recent face datasets have been collected from the web \cite{DBLP:journals/corr/abs-2102-00813} which raises the question of consent of the individuals depicted in these images. However, due to legal and privacy issues \cite{gdpr, DBLP:conf/iccv/QiuYG00T21,DBLP:journals/tifs/MedenRTDKSRPS21} the use of face datasets collected from the web might not be possible in the future anymore. Privacy regulations such as the GDPR \cite{gdpr} assure individuals the right to withdraw their consent to use or store their private data, practically making the use and distribution of large face datasets impossible. These circumstances call for a solution that considers the privacy of individuals, which synthetically generated images can support. A recent work followed this motivation to take advantage of synthetic data to develop MADs in a privacy-friendly frame \cite{DBLP:journals/corr/abs-2203-06691}.

Driven by the legal and ethical concerns of using real images to develop MADs, and the need for well-generalized MADs developed on the basis of large and diverse datasets, we conducted the SYN-MAD 2022: Competition on Face Morphing Attack Detection Based on Privacy-aware Synthetic Training Data at the International Joint Conference on Biometrics 2022. The results and observations are summarized in this paper.

\vspace{-2.5mm}
\section{Datasets, Evaluation Criteria, and Participants} 
\vspace{-1.5mm}
\subsection{Provided Training Dataset}
\vspace{-1.5mm}
All participants were provided with the training set of the Synthetic Morphing Attack Detection Development (SMDD) dataset \cite{DBLP:journals/corr/abs-2203-06691}, and all solutions were allowed to use only this dataset to train the face morphing attack detector. The SMDD dataset was created using the official open-source implementation of StyleGAN2-ADA \cite{DBLP:conf/nips/KarrasAHLLA20}. As detailed in \cite{DBLP:journals/corr/abs-2203-06691}, first, 500k images were generated using a random Gaussian noise vector drown from a normal distribution. These images were then randomly divided for training and test set, and then 50k images were selected based on their calculated face image quality. The face image quality was determined using CR-FIQA \cite{DBLP:journals/corr/abs-2112-06592} and describes the utility of the images for face recognition, as defined in ISO/IEC 29794-1 \cite{ISOIEC29794-1}. The selected 50k images were again randomly split into two parts of equal size. The first part was considered the bona fide samples and was not used for morphing, while the second part was used to create the morphed images. 5k images of the second part were randomly chosen as the key morphing images and were mated with five randomly chosen images from the second part. For the morphing of the training images, the OpenCV/dlib morphing algorithm \cite{openCVmorph} is used for each pair.

In total, the training split of the SMDD dataset consists of 25k bona fide and 15k morphing images based on synthetic faces, which avoids using privacy-sensitive real face images. Furthermore, the bounding box and five facial landmark points of all the images were additionally provided to the participants.

\vspace{-1.8mm}
\subsection{Evaluation Benchmark} 
\vspace{-1.5mm}
For the evaluation, a new MAD evaluation benchmark, MAD22, was created by the organizers as part of the competition and will be publicly released\footnote{https://github.com/marcohuber/SYN-MAD-2022}. This dataset is based on the images of the Face Research Lab London (FRLL) dataset \cite{DeBruine2021}. The FRLL dataset contains images of 102 different identities and provides high-quality close-up frontal face images created under uniform illumination with a wide range of ethnicities. All individuals in the dataset signed consent for their images to be "used in lab-based and web-based studies in their original or altered forms and to illustrate research (e.g., in scientific journals, news media or presentations)." The dataset itself has been released under the CC BY 4.0 license. 
For the morph generation, we limit the data to the frontal images of the dataset, consisting of "neutral\_front" and "smiling\_front" as most morphing approaches are optimized for frontal images. 

\vspace{-4mm}
\subsubsection{Selection of the Morphing Pairs}
\vspace{-2mm}
For selecting the morphing pairs, we split the frontal images of FRLL depending on the provided gender label and retained the neutral and smiling split. We utilize ElasticFace-Arc \cite{DBLP:journals/corr/abs-2109-09416} to generate embeddings of all the images and compare all possible pairs with cosine similarity. The pairs are then ordered depending on their achieved similarity score, and we select the 250 most similar pairs as the morphing pairs to obtain challenging, realistic morphs. Using the splits, we get a total of 1000 morphing pairs: 250 female neutral pairs, 250 female smiling pairs, 250 male neutral pairs, and 250 male smiling pairs.

\vspace{-4mm}
\subsubsection{Morphing Approaches}
\vspace{-2mm}
\label{morphing}
Starting from the 1000 image pairs selected by their similarity, we utilized five different morphing approaches, three landmark-based approaches, and two generative adversarial network (GAN)-based approaches to ensure diversity in the attacks. 

The \textbf{OpenCV} algorithm is based on the "Face Morph Using OpenCV" tutorial \footnote{https://learnopencv.com/face-morph-using-opencv-cpp-python/} and uses the Dlib \cite{DBLP:journals/jmlr/King09} implementation of the landmark detector proposed by Kazemi and Sullivan \cite{DBLP:conf/cvpr/KazemiS14}. The detected landmarks are used to perform Delaunay triangulation on each image, and based on affine transformation, the triangles are wrapped and blended. 16 out of the 1000 morph images showed strong morphing artifacts (a black region on the mouth area). These images are removed, resulting in 984 morph images based on the OpenCV algorithm. The same morphing approach has also been used in the development dataset to create the morphs of the synthetic training dataset, the SMDD.

The \textbf{FaceMorpher} is also a landmark-based morphing tool. It is a commercial-of-the-shelf (COTS) face morphing tool. More details on the technical aspects of the approach are not released.

\textbf{Webmorph} is a landmark-based online tool\footnote{https://webmorph.org/} optimized for averaging and transforming faces. It can also be used to create morphs of two images. As the program often failed to create morphs without huge artifacts for images with smiling, we only created morphs of the neutral frontal images, resulting in just 500 morphed images based on the Webmorph face morphing approach.

For approaches based on GANs \cite{DBLP:conf/cvpr/KarrasLA19}, we selected \textbf{MIPGAN-I} and \textbf{MIPGAN-II} \cite{DBLP:journals/tbbis/ZhangVRRDB21}. MIPGAN utilizes a loss function that incorporates perceptual quality, identity, identity differences and structural visibility into StyleGAN, either StyleGAN \cite{DBLP:conf/cvpr/KarrasLA19} (MIPGAN-1) or StyleGAN2 \cite{DBLP:conf/cvpr/KarrasLAHLA20} (MIPGAN-2). Both approaches produce high-resolution morphs with minimal artifacts. In one case, the detection of facial landmarks failed on a morphed image, which reduces the amount of morphs created with the MIPGAN-2 approach to 999.

\begin{figure*}
    \centering
    \includegraphics[width=0.85\textwidth]{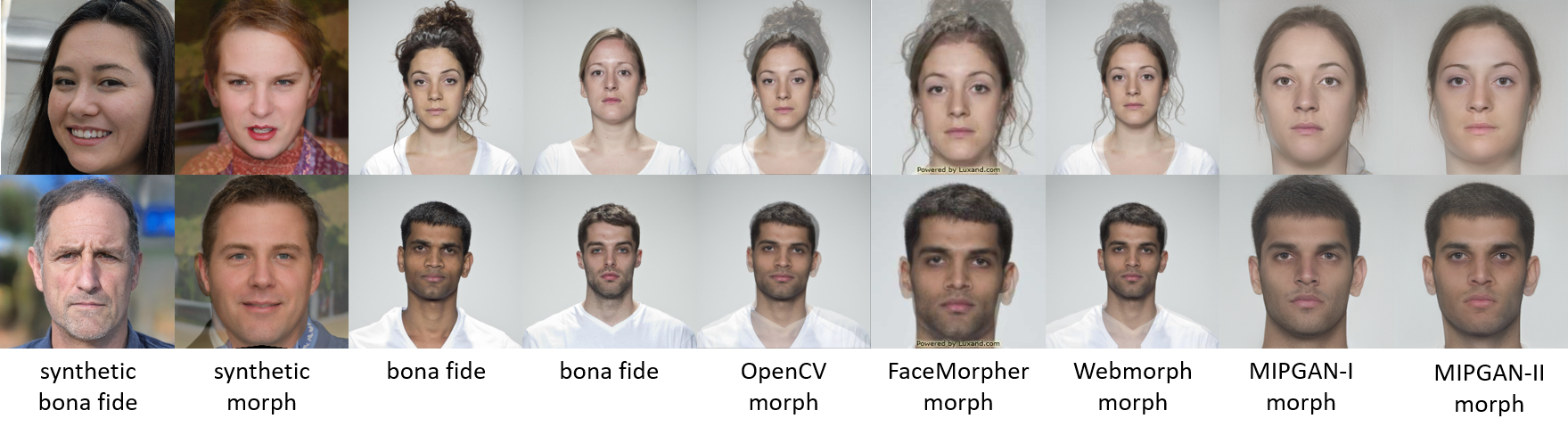}
    \caption{Examples of the used images during the competition: The synthetic bona fide and synthetic morphs were provided to the competition participants and are part of the training split of the SMDD dataset \cite{DBLP:journals/corr/abs-2203-06691}. The bona fide images are taken from the FRLL dataset \cite{DeBruine2021} and they also serve as the images used for the different morphing approaches to create our MAD22 benchmark. The evaluation morphs were created by the competition organizers and will be released to the public.}
    \label{examples}
\end{figure*}

The created MAD22 benchmark contains 4483 (984 OpenCV, 1000 FaceMorpher, 500 Webmorph, 1000 MIPGAN-I, 999 MIPGAN-II) morphed face images and 204 bona fide images from the FRLL dataset. Some example images of the synthetic training dataset, as well as the created MAD22 benchmark, are shown in Figure \ref{examples}.

\vspace{-4mm}
\subsubsection{Baseline Approach}
\vspace{-2mm}
The baseline performance evaluation is based on MixFaceNet \cite{DBLP:journals/corr/abs-2203-06691, DBLP:conf/icb/BoutrosDFKK21} It has shown good performances in face presentation attack and morphing attack detection and possesses a low computational complexity while retaining high accuracy. The architecture also takes advantage of convolutional kernels of different sizes to capture different levels of face attack clues \cite{DBLP:conf/fgr/FangBKD21}.

\vspace{-1mm}
\subsection{Evaluation Criteria} 
\vspace{-1mm}
For the vulnerability analysis, we look at the False Non-Match Rate (FNMR) at two different, fixed False match rates (FMR), FMR100 and FMR1000, to show the strength of the attacks. The FMR100 refers to the operational point that achieves the lowest FNMR with an FMR $<$ 1.0\% and the FMR1000 to the operational point with FMR $<$ 0.1\%. To investigate the vulnerability of the used face recognition models, we calculate the Mated Morph Presentation Match Rate (MMPMR) \cite{DBLP:conf/biosig/ScherhagNRGVSSM17}. The MMPMR refers to the fraction of morphs whose similarity to both identities used to morph, are below the selected threshold relative to all morphs. In other words, the proportion of morphs that match both identities used to morph out of all morphs.

The evaluation of the morphing attack detector performance will be based on the ISO/IEC 30107-3 \cite{ISO301073} standard and, therefore, presented as the Bona fide Presentation Classification Error Rate (BPCER) and the Attack presentation Classification Error Rate (APCER). The BPCER refers to the proportion of bona fide images incorrectly classified as attack samples, and the APCER refers to the proportion of attack images incorrectly classified as bona fide samples. Furthermore, we will report the APCER at a fixed BPCER to provide a more detailed analysis of the different approaches and their performance. To cover different operational points and to present comparative results, the submitted solutions are evaluated at three different fixed APCER (BPCER) values, 0.1\%, 1.0\%, 10\%, and 20\%, and the corresponding BPCER (APCER) is reported. The final ranking on the submitted solutions is based on the APCER at BPCER of 20\%.
As the bona fide samples are the same for all attack types, fixing the BPCER would produce the same operational threshold. This will allow us to analyze the detectability of the different attacks at the same operation threshold, and thus the APCER at fixed BPCER is considered for the most of the discussion and the final ranking.

\vspace{-1.3mm}
\subsection{Vulnerability Analysis} 
\vspace{-1.3mm}
To investigate the success of the created morphing attacks, we performed a vulnerability analysis using two different state-of-the-art face recognition models, CurricularFace \cite{DBLP:conf/cvpr/HuangWT0SLLH20} and ElasticFace-Arc \cite{DBLP:journals/corr/abs-2109-09416} \footnote{CurricularFace and ElasticFace utilized ResNet100 as backbone trained on MS1MV dataset}. To show the suitability of the chosen face recognition models, we also report their performance in terms of FNMR@FMR (in percentage) on the Labeled Faces in the Wild (LFW) \cite{LFWTech} benchmark. On LFW, CurricularFace \cite{DBLP:conf/cvpr/HuangWT0SLLH20} achieved 0.3\% at FMR100 and 0.33\% at FMR1000. ElasticFace-Arc, in comparison, achieved 0.23\% at FMR100 and 0.3\% at FMR1000. Both models achieved top performance on mainstream benchmarks as reported in \cite{DBLP:journals/corr/abs-2109-09416}. The FMR1000 and FMR100 decision thresholds used in the vulnerability study were calculated on the LFW benchmark. For the analysis of the vulnerability, we take the morphed image $M_{A_{n}, B_{n}}$ of two images from two identities from one scenario (e.g. neutral), $ID_{A_{n}}$ and $ID_{B_{n}}$ and compare them to their counterpart in the other scenario (e.g. smiling), $ID_{A_{S}}$ and $ID_{B_{s}}$. If the similarity of the morphed image  $M_{A_{n}, B_{n}}$ to the original images $ID_{A_{S}}$ and $ID_{B_{s}}$ is higher than the set threshold, the system is considered vulnerable to this attack. The results of the vulnerability analysis are summarized in Table \ref{vuln}. The results show the vulnerability (MMPMR) of the face recognition models to the morphed images, as in most cases, over 90\% of the morphed images are matched with both identities used to create the morphed image, even at high similarity thresholds. 

\begin{table}[]
\centering
\small
\resizebox{0.40\textwidth}{!}{
\begin{tabular}{cc|cc}
\hline
\multicolumn{1}{l}{Morphing Approach} & \multicolumn{1}{l|}{FR Model} & $T_{FMR100}$ & $T_{FMR1000}$ \\ \hline\hline
\multirow{2}{*}{OpenCV}               & CF                              & 0.996        & 0.986         \\
                                      & EF-Arc                             & 0.997        & 0.980         \\ \hline
\multirow{2}{*}{FaceMorpher}          & CF                             & 0.970        & 0.935         \\
                                      & EF-Arc                             & 0.962        & 0.913         \\ \hline
\multirow{2}{*}{Webmorph}             & CF                            & 0.988        & 0.988         \\
                                      & EF-Arc                             & 0.990        & 0.986         \\ \hline
\multirow{2}{*}{MIPGAN-I}             & CF                             & 0.962        & 0.890         \\
                                      & EF-Arc                             & 0.980        & 0.845         \\ \hline
\multirow{2}{*}{MIPGAN-II}            & CF                             & 0.953        & 0.832         \\
                                      & EF-Arc                             & 0.953        & 0.778         \\ \hline
\end{tabular}}
\vspace{0.3mm}
\caption{Vulnerability analysis in terms of MMPMR at two different operational points (FMR100 and FMR1000) on two face recognition (FR) models, CurricularFace (CF) and ElasticFace-Arc (EF-Arc). The results show that both face recognition models are vulnerable to the different morphing approaches as in most cases, even with high thresholds in terms of required similarity to be matched, the morphed images are matched with both identities used to create the morph in over 90\% of the cases.}
\vspace{-4.5mm}
\label{vuln}
\end{table}

\vspace{-1.3mm}
\subsection{Competition Participants} 
\vspace{-1.3mm}
The competition aimed at attracting participants, both from academia and industry, with a high geographic and activity variation. The call for participation was shared on the International Joint Conference on Biometrics 2022 website, on the competition's own website, on several social media platforms, and through private e-mailing lists. The call for participation has attracted 12 registered teams from both academia and industry. Out of these, six teams submitted a valid solution. These six teams have affiliations in four different countries. Three teams have academic affiliations, two teams have industry affiliations, and one team has mixed, both academic and industry backgrounds. Only one team has chosen to be anonymous, and one team submitted two solutions as each team was allowed to submit up to two solutions. The total number of validly submitted solutions is 7. A summary of the participating teams is presented in Table \ref{teams}.

\vspace{-2mm}
\subsection{Submission and Evaluation Process}
\vspace{-2mm}
Each team had to request the synthetic data first and provide their team name and affiliation. Only the synthetic data, consisting of the synthetic morphs and synthetic bona fide images as well as text files containing information about the bounding box and facial landmarks for each image, has been provided. Each team was then requested to provide a Win32 or a Linux executable or, if wanted, provide their python script. The teams had to provide an executable to evaluate their approach on the classified evaluation benchmark and also an executable to re-train their model. The re-training executable is to be used by the organizing team to validate that solely the provided synthetic data is used to train the approach, and no other pre-trained weights of any kind are used. All the approaches have been evaluated on a restricted system without an internet connection to prevent any data leak. At no time the participating teams had any access or knowledge about the evaluation benchmark used during the competition, including any information about the morphing approaches used or the images source dataset.

\vspace{-3mm}
\section{Submitted Solutions}
\vspace{-2.5mm}
\label{solutions}
In total, 12 teams have registered for the competition. Each team was allowed to submit up to two submissions. Eventually, seven valid submissions from six different teams were received. Solution names, team members, affiliations, and type of the institution, i.e., academic, industry, or mixed, are summarized in Table \ref{teams}. One team opted to keep their names and affiliations anonymized. A condensed summary of the details of the approaches (e.g., input size, architecture, loss function, etc.) is listed in Table \ref{approaches}. In the following, we provide a brief description of the valid submitted solutions:


\begin{table*}[]
\centering
\small
\begin{tabular}{llll} \hline 
Solution           & Team Members                              & Affiliations                                 & Type     \\ \hline \hline
Xception           & Borut Batagelj, Andrej Kronov\v{s}ek,     & Faculty of Computer and Information Science, & Academic \\
                   & Peter Peer                                & University of Ljubljana                      &          \\ \hline
MorphHRNet         & Marija Ivanovska, Borut Batagelj,         & Faculty of Electrical Engineering,           & Academic \\
                   & Peter Peer, Vitomir \v{S}truc             & Faculty of Computer and Information Science, &          \\ 
                   &                                           & University of Ljubljana                      &          \\ \hline
E-CBAM@VCMI        & Pedro C. Neto, Tiago Gon\c{c}alves,       & INESC TEC, Faculty of Engineering,           & Academic \\
                   & Ana F. Sequeira, Jaime S. Cardoso         & University of Porto                          &          \\ \hline
Con-Text Net (A/B) & Jo\~{a}o Tremo\c{c}o, Miguel Louren\c{c}o & Yoonik                                       & Industry \\ \hline
VaMoS              & Sergio Serra, Eduardo Cerme\~{n}o         & Vaelsys R\&D                                 & Industry \\ \hline
Anonymous          & Anonymous                                 & Anonymous                                    & mixed    \\ \hline
\end{tabular}
\vspace{0.5mm}
\caption{A summary of the valid submitted solutions, participating team members, affiliations, and type of institution. More details on the submitted algorithms are provided in Section 3.}
\label{teams}
\end{table*}

\begin{table*}[]
\centering
\small
\begin{tabular}{llllll} \hline 
Solution       & Input size           & Architecture/    & Loss function    & Evaluation data           & Bounding box/   \\
               & (in pixels)          & Model            &                  & during training           & Landmarks?      \\ \hline \hline
MorphHRNet     & $256 \times 256$     & HRNet            & Binary cross     & 20\% of provided data,    & bounding box    \\
               &                      &                  & entropy          & FRLL, FERET, FRGC morphs  &                 \\ \hline
Con-Text Net A & $128 \times 128$     & 2xResNet-50      & Weighted Focal   & New Face Morphing Dataset & landmarks       \\
               & and $256 \times 256$ &                  &                  &                           &                 \\ \hline
Xception       & $299 \times 299$     & Xception net     & Cross-entropy    & FRLL, FERET,  FRGC morphs  & other alignment \\ \hline
Con-Text Net B & $128 \times 128$     & 2xResNet-50      & Weighted Focal   & New Face Morphing Dataset & landmarks       \\ 
               & and $256 \times 256$ &                  &                  &                           &                 \\ \hline
E-CBAM@VCMI    & $256 \times 256$     & 4xResNet-50      & Cross-entropy    & 25\% of the data for      & no alignment    \\
               &                      & with CBAM        &                  & different models          &                 \\ \hline
Anonymous      & $256 \times 256$     & Hand-crafted+SVM & ---              & no other data             & bounding box    \\ \hline
VaMoS          & $112 \times 112$     & ResNet-50        & Additive Angular & only provided data        & bounding box +  \\
               &                      &                  & Margin           &                           & landmarks       \\ \hline          
\end{tabular}
\vspace{0.4mm}
\caption{Basic details of the submitted approaches. ResNet-50 is the most common utilized model architecture in the competition and the used input size varies. More details on the submitted algorithms are provided in Section 3.}
\label{approaches}
\end{table*}


\textbf{Xception}\footnote{\label{note1}The research was supported in parts by the Slovenian Research Agency (ARRS) in the scope of the research project J2-1734 (B) "FaceGEN"}: This approach utilizes the Xception-net architecture  \cite{DBLP:conf/cvpr/Chollet17} modified by an additional 2-class classification layer on top of the network. The choice of Xception network is based on its high performance in detecting DeepFakes \cite{DBLP:conf/iccv/RosslerCVRTN19}. The approach is trained for 39 epochs with 2500 iterations per epoch utilizing Adam optimizer \cite{DBLP:journals/corr/KingmaB14}. The learning rate is set to 0.0001, and a weight decay to 0.001 is applied. Instead of using the provided landmarks and bounding boxes, face detection is done based on RetinaFace detector \cite{DBLP:conf/cvpr/DengGVKZ20}. During the development, the approach has been evaluated on FRLL morphs, FERET morphs, and FRGC morphs \cite{DBLP:journals/corr/abs-2012-05344}. The input to the model is a 40\% enlarged frame of the found face. The face image is then scaled to $299 \times 299$ pixels and randomly flipped horizontally. All images are then also normalized across all channels. More details on the approach are presented in \cite{batagelj2022morphing}.

\textbf{MorphHRNet}$^{5}$: This approach is based on training an HRNet \cite{DBLP:conf/cvpr/0009XLW19, DBLP:journals/pami/00010CJDZ0MTW0X21} to perform two-class classification i.e, bona fide or morphing attack. The approach is trained using binary cross-entropy and optimized using the Adam optimizer \cite{DBLP:journals/corr/KingmaB14}. During training, the model is evaluated on a combination of the FRLL morphs, FERET morphs, and FRGC morphs \cite{DBLP:journals/corr/abs-2012-05344} and also 20\% of the provided synthetic data. The best-performing model based on the validation dataset is selected. The model is trained for 20 epochs with a learning rate of 0.001. Furthermore, the provided bounding boxes have been used during pre-processing. As pre-processing, the images have been resized and normalized. For data augmentation, horizontal flipping and random rotations up to 5 degrees have been applied. More details on the approach are presented in \cite{batagelj2022morphing}.

\begin{table*}[]
\centering
\small
\resizebox{0.85\textwidth}{!}{
\begin{tabular}{cllllllllll}
\multicolumn{1}{l}{} &  &  &  &  &  &  &  &  &  &  \\ \cline{2-11} 
\multicolumn{1}{l|}{} & \multicolumn{1}{c|}{\begin{tabular}[c]{@{}c@{}}Morphing\\ approach\end{tabular}} & \multicolumn{9}{c|}{OpenCV} \\ \hline
\multicolumn{1}{|c|}{Rank} & \multicolumn{1}{c|}{Solution} & \multicolumn{1}{c|}{\textbf{\begin{tabular}[c]{@{}c@{}}APCER@\\ BPCER\\ 20\%\end{tabular}}} & \multicolumn{1}{c|}{\begin{tabular}[c]{@{}c@{}}APCER@\\ BPCER\\ 10\%\end{tabular}} & \multicolumn{1}{c|}{\begin{tabular}[c]{@{}c@{}}APCER@\\ BPCER\\ 1\%\end{tabular}} & \multicolumn{1}{c|}{} & \multicolumn{1}{c|}{\begin{tabular}[c]{@{}c@{}}BPCER@\\ APCER\\ 20\%\end{tabular}} & \multicolumn{1}{c|}{\begin{tabular}[c]{@{}c@{}}BPCER@\\ APCER\\ 10\%\end{tabular}} & \multicolumn{1}{c|}{\begin{tabular}[c]{@{}c@{}}BPCER@\\ APCER\\ 1\%\end{tabular}} & \multicolumn{1}{l|}{} & \multicolumn{1}{c|}{EER} \\ \cline{1-5} \cline{7-9} \cline{11-11} 
\multicolumn{1}{|c|}{-} & \multicolumn{1}{l|}{Baseline} & \multicolumn{1}{l|}{0.0376} & \multicolumn{1}{l|}{0.065} & \multicolumn{1}{l|}{0.3638} & \multicolumn{1}{l|}{} & \multicolumn{1}{l|}{0.0345} & \multicolumn{1}{l|}{0.0739} & \multicolumn{1}{l|}{0.532} & \multicolumn{1}{l|}{} & \multicolumn{1}{l|}{0.0833} \\ \cline{1-5} \cline{7-9} \cline{11-11} 
\multicolumn{1}{|c|}{1} & \multicolumn{1}{l|}{MorphHRNet} & \multicolumn{1}{l|}{\textbf{0.0163}} & \multicolumn{1}{l|}{0.0376} & \multicolumn{1}{l|}{0.6697} & \multicolumn{1}{l|}{} & \multicolumn{1}{l|}{0.0147} & \multicolumn{1}{l|}{0.0196} & \multicolumn{1}{l|}{0.3382} & \multicolumn{1}{l|}{} & \multicolumn{1}{l|}{0.0569} \\ \cline{1-5} \cline{7-9} \cline{11-11} 
\multicolumn{1}{|c|}{2} & \multicolumn{1}{l|}{Xception} & \multicolumn{1}{l|}{0.0254} & \multicolumn{1}{l|}{0.0661} & \multicolumn{1}{l|}{0.2175} & \multicolumn{1}{l|}{} & \multicolumn{1}{l|}{0.0147} & \multicolumn{1}{l|}{0.049} & \multicolumn{1}{l|}{0.3529} & \multicolumn{1}{l|}{} & \multicolumn{1}{l|}{0.0732} \\ \cline{1-5} \cline{7-9} \cline{11-11} 
\multicolumn{1}{|c|}{3} & \multicolumn{1}{l|}{Con-Text Net A} & \multicolumn{1}{l|}{0.1575} & \multicolumn{1}{l|}{0.2652} & \multicolumn{1}{l|}{0.7093} & \multicolumn{1}{l|}{} & \multicolumn{1}{l|}{0.1618} & \multicolumn{1}{l|}{0.3284} & \multicolumn{1}{l|}{0.7402} & \multicolumn{1}{l|}{} & \multicolumn{1}{l|}{0.1748} \\ \cline{1-5} \cline{7-9} \cline{11-11} 
\multicolumn{1}{|c|}{4} & \multicolumn{1}{l|}{Con-Text Net B} & \multicolumn{1}{l|}{0.2368} & \multicolumn{1}{l|}{0.3445} & \multicolumn{1}{l|}{0.5783} & \multicolumn{1}{l|}{} & \multicolumn{1}{l|}{0.2353} & \multicolumn{1}{l|}{0.4314} & \multicolumn{1}{l|}{0.8235} & \multicolumn{1}{l|}{} & \multicolumn{1}{l|}{0.2266} \\ \cline{1-5} \cline{7-9} \cline{11-11} 
\multicolumn{1}{|c|}{5} & \multicolumn{1}{l|}{E-CBAM@VCMI} & \multicolumn{1}{l|}{0.3303} & \multicolumn{1}{l|}{0.4868} & \multicolumn{1}{l|}{0.9858} & \multicolumn{1}{l|}{} & \multicolumn{1}{l|}{0.3824} & \multicolumn{1}{l|}{0.5049} & \multicolumn{1}{l|}{0.7647} & \multicolumn{1}{l|}{} & \multicolumn{1}{l|}{0.2754} \\ \cline{1-5} \cline{7-9} \cline{11-11} 
\multicolumn{1}{|c|}{6} & \multicolumn{1}{l|}{Anonymous} & \multicolumn{1}{l|}{0.4461} & \multicolumn{1}{l|}{0.999} & \multicolumn{1}{l|}{1.000} & \multicolumn{1}{l|}{} & \multicolumn{1}{l|}{0.4363} & \multicolumn{1}{l|}{0.5049} & \multicolumn{1}{l|}{0.6078} & \multicolumn{1}{l|}{} & \multicolumn{1}{l|}{0.3252} \\ \cline{1-5} \cline{7-9} \cline{11-11} 
\multicolumn{1}{|c|}{7} & \multicolumn{1}{l|}{VaMoS} & \multicolumn{1}{l|}{--} & \multicolumn{1}{l|}{--} & \multicolumn{1}{l|}{--} & \multicolumn{1}{l|}{} & \multicolumn{1}{l|}{0.9118} & \multicolumn{1}{l|}{0.9853} & \multicolumn{1}{l|}{1.000} & \multicolumn{1}{l|}{} & \multicolumn{1}{l|}{--} \\ \cline{1-5} \cline{7-9} \cline{11-11} 
\end{tabular}}
\vspace{0.3mm}
\caption{The evaluation results on the landmark-based morph dataset utilizing the OpenCV morphing approach. The same morphing approach was used to create the synthetic morphs of the training dataset.}
\label{opencv}
\end{table*}

\begin{table*}[]
\centering
\small
\resizebox{0.85\textwidth}{!}{
\begin{tabular}{cllllllllll}
\multicolumn{1}{l}{} &  &  &  &  &  &  &  &  &  &  \\ \cline{2-11} 
\multicolumn{1}{l|}{} & \multicolumn{1}{c|}{\begin{tabular}[c]{@{}c@{}}Morphing\\ approach\end{tabular}} & \multicolumn{9}{c|}{FaceMorpher} \\ \hline
\multicolumn{1}{|c|}{Rank} & \multicolumn{1}{c|}{Solution} & \multicolumn{1}{c|}{\textbf{\begin{tabular}[c]{@{}c@{}}APCER@\\ BPCER\\ 20\%\end{tabular}}} & \multicolumn{1}{c|}{\begin{tabular}[c]{@{}c@{}}APCER@\\ BPCER\\ 10\%\end{tabular}} & \multicolumn{1}{c|}{\begin{tabular}[c]{@{}c@{}}APCER@\\ BPCER\\ 1\%\end{tabular}} & \multicolumn{1}{c|}{} & \multicolumn{1}{c|}{\begin{tabular}[c]{@{}c@{}}BPCER@\\ APCER\\ 20\%\end{tabular}} & \multicolumn{1}{c|}{\begin{tabular}[c]{@{}c@{}}BPCER@\\ APCER\\ 10\%\end{tabular}} & \multicolumn{1}{c|}{\begin{tabular}[c]{@{}c@{}}BPCER@\\ APCER\\ 1\%\end{tabular}} & \multicolumn{1}{l|}{} & \multicolumn{1}{c|}{EER} \\ \cline{1-5} \cline{7-9} \cline{11-11} 
\multicolumn{1}{|c|}{-} & \multicolumn{1}{l|}{Baseline} & \multicolumn{1}{l|}{0.029} & \multicolumn{1}{l|}{0.036} & \multicolumn{1}{l|}{0.055} & \multicolumn{1}{l|}{} & \multicolumn{1}{l|}{0.0049} & \multicolumn{1}{l|}{0.0049} & \multicolumn{1}{l|}{1.000} & \multicolumn{1}{l|}{} & \multicolumn{1}{l|}{0.046} \\ \cline{1-5} \cline{7-9} \cline{11-11} 
\multicolumn{1}{|c|}{1} & \multicolumn{1}{l|}{Con-Text Net B} & \multicolumn{1}{l|}{\textbf{0.000}} & \multicolumn{1}{l|}{0.000} & \multicolumn{1}{l|}{0.000} & \multicolumn{1}{l|}{} & \multicolumn{1}{l|}{0.000} & \multicolumn{1}{l|}{0.000} & \multicolumn{1}{l|}{0.000} & \multicolumn{1}{l|}{} & \multicolumn{1}{l|}{0.000} \\ \cline{1-5} \cline{7-9} \cline{11-11} 
\multicolumn{1}{|c|}{2} & \multicolumn{1}{l|}{Xception} & \multicolumn{1}{l|}{0.000} & \multicolumn{1}{l|}{0.000} & \multicolumn{1}{l|}{0.005} & \multicolumn{1}{l|}{} & \multicolumn{1}{l|}{0.0147} & \multicolumn{1}{l|}{0.0147} & \multicolumn{1}{l|}{0.0147} & \multicolumn{1}{l|}{} & \multicolumn{1}{l|}{0.006} \\ \cline{1-5} \cline{7-9} \cline{11-11} 
\multicolumn{1}{|c|}{3} & \multicolumn{1}{l|}{Con-Text Net A} & \multicolumn{1}{l|}{0.000} & \multicolumn{1}{l|}{0.000} & \multicolumn{1}{l|}{0.999} & \multicolumn{1}{l|}{} & \multicolumn{1}{l|}{0.000} & \multicolumn{1}{l|}{0.000} & \multicolumn{1}{l|}{1.000} & \multicolumn{1}{l|}{} & \multicolumn{1}{l|}{0.000} \\ \cline{1-5} \cline{7-9} \cline{11-11} 
\multicolumn{1}{|c|}{4} & \multicolumn{1}{l|}{MorphHRNet} & \multicolumn{1}{l|}{0.024} & \multicolumn{1}{l|}{0.043} & \multicolumn{1}{l|}{0.312} & \multicolumn{1}{l|}{} & \multicolumn{1}{l|}{0.0147} & \multicolumn{1}{l|}{0.0196} & \multicolumn{1}{l|}{0.4804} & \multicolumn{1}{l|}{} & \multicolumn{1}{l|}{0.059} \\ \cline{1-5} \cline{7-9} \cline{11-11} 
\multicolumn{1}{|c|}{5} & \multicolumn{1}{l|}{E-CBAM@VCMI} & \multicolumn{1}{l|}{0.628} & \multicolumn{1}{l|}{0.928} & \multicolumn{1}{l|}{1.000} & \multicolumn{1}{l|}{} & \multicolumn{1}{l|}{0.8235} & \multicolumn{1}{l|}{0.951} & \multicolumn{1}{l|}{1.000} & \multicolumn{1}{l|}{} & \multicolumn{1}{l|}{0.412} \\ \cline{1-5} \cline{7-9} \cline{11-11} 
\multicolumn{1}{|c|}{6} & \multicolumn{1}{l|}{Anonymous} & \multicolumn{1}{l|}{0.999} & \multicolumn{1}{l|}{0.999} & \multicolumn{1}{l|}{1.000} & \multicolumn{1}{l|}{} & \multicolumn{1}{l|}{0.7255} & \multicolumn{1}{l|}{0.7404} & \multicolumn{1}{l|}{1.000} & \multicolumn{1}{l|}{} & \multicolumn{1}{l|}{0.653} \\ \cline{1-5} \cline{7-9} \cline{11-11} 
\multicolumn{1}{|c|}{7} & \multicolumn{1}{l|}{VaMoS} & \multicolumn{1}{l|}{--} & \multicolumn{1}{l|}{--} & \multicolumn{1}{l|}{--} & \multicolumn{1}{l|}{} & \multicolumn{1}{l|}{0.9069} & \multicolumn{1}{l|}{0.9804} & \multicolumn{1}{l|}{1.000} & \multicolumn{1}{l|}{} & \multicolumn{1}{l|}{--} \\ \cline{1-5} \cline{7-9} \cline{11-11} 
\end{tabular}}
\vspace{0.3mm}
\caption{The evaluation results on the landmark-based morphs created with the COTS system FaceMorpher. When investigating the details, it is noticeable that some systems find it very easy to recognize the morphs, resulting in no errors even at very restrictive thresholds.}
\label{facemorpher}
\end{table*}

\textbf{Anonymous:} The face morphing detector exploits the idea introduced by \cite{DBLP:conf/iwdw/Neubert17}. It is based on the idea of the change caused by JPEG compression in morphed images in comparison to bona fide images. In total, 46 features are extracted from the original images and the compressed image. On these extracted features, a support vector machine (SVM) is trained with the SMO solver. During the feature creation process, the provided bounding boxes are utilized. The team emphasizes that the approach will only work on very similar images to the synthetic training images, and no effort has been made to develop a generalizable approach.

\textbf{E-CBAM@VCMI}: This approach is based on training ResNet-50 architecture  \cite{DBLP:conf/cvpr/HeZRS16} with attention module \cite{DBLP:conf/eccv/WooPLK18} on the provided dataset. The provided dataset is divided into 4 splits with 25\% of the data without overlapping. Four different models are then trained on 75\% of the provided training dataset and are validated on a different split. For the evaluation, all the images are processed by the four different models for the score computation, and the mean value is taken as the final score. The models are optimized using cross-entropy loss and are trained for 20 epochs each. The learning rate was set to 0.0001. No bounding boxes or alignment has been used. Besides center cropping the images to $256 \times 256$, along with normalization, random affine transformation, and random horizontal splits have been performed. 

\textbf{Con-Text Net A/B:} The approach consisted of two ResNet-50 backbones \cite{DBLP:conf/cvpr/HeZRS16}. One backbone is trained using input images cropped to the size $128 \times 128$ using the provided landmarks. The other backbone is trained using the original image resized to $256 \times 256$. The linearized outputs of each backbone are 2048-dimensional feature vectors than are then fed to a classifier with three fully connected layers with SiLU activations \cite{DBLP:journals/corr/ElfwingUD17}. Both models are trained with a batch size of 24 with Adam optimizer \cite{DBLP:journals/corr/KingmaB14}. The learning rate was initially set to 0.01 and decreased with an exponential learning rate scheduler with factor of 0.85. The utilized learning objective was weighted focal loss. During development, the Biometix "New Face Morphing Dataset (for vulnerability research)" \footnote{https://www.linkedin.com/pulse/new-face-morphing-dataset-vulnerability-research-ted-dunstone/} has been utilized to evaluate the approach. The Con-Text Net A has been trained for 18 epochs, and the Con-Text Net B has been trained for 22 epochs.

\textbf{VaMoS:} This approach is based on previous works on the concealment of morphing attacks \cite{DBLP:journals/access/CarabeC21} and solutions to prevent registration using fake images \cite{carabemethods}. In comparison to the previous mentioned works, this work utilized ArcFace loss \cite{DBLP:conf/cvpr/DengGXZ19} instead of inception net trained with triplet loss (FaceNet) \cite{DBLP:conf/cvpr/SchroffKP15}. Furthermore, the neural network architecture has been modified to not longer require a morphing detector \cite{DBLP:conf/icb/RaghavendraRVB17}. The model architecture is based on the ResNet-50 \cite{DBLP:conf/cvpr/HeZRS16} architecture and is trained for 100 epochs with a learning rate of 0.001 with ArcFace loss \cite{DBLP:conf/cvpr/DengGXZ19}. The approach utilizes the provided bounding boxes and face landmarks.

\begin{table*}[]
\centering
\small
\resizebox{0.85\textwidth}{!}{
\begin{tabular}{cllllllllll}
\multicolumn{1}{l}{} &  &  &  &  &  &  &  &  &  &  \\ \cline{2-11} 
\multicolumn{1}{l|}{} & \multicolumn{1}{c|}{\begin{tabular}[c]{@{}c@{}}Morphing\\ approach\end{tabular}} & \multicolumn{9}{c|}{Webmorph} \\ \hline
\multicolumn{1}{|c|}{Rank} & \multicolumn{1}{c|}{Solution} & \multicolumn{1}{c|}{\textbf{\begin{tabular}[c]{@{}c@{}}APCER@\\ BPCER\\ 20\%\end{tabular}}} & \multicolumn{1}{c|}{\begin{tabular}[c]{@{}c@{}}APCER@\\ BPCER\\ 10\%\end{tabular}} & \multicolumn{1}{c|}{\begin{tabular}[c]{@{}c@{}}APCER@\\ BPCER\\ 1\%\end{tabular}} & \multicolumn{1}{c|}{} & \multicolumn{1}{c|}{\begin{tabular}[c]{@{}c@{}}BPCER@\\ APCER\\ 20\%\end{tabular}} & \multicolumn{1}{c|}{\begin{tabular}[c]{@{}c@{}}BPCER@\\ APCER\\ 10\%\end{tabular}} & \multicolumn{1}{c|}{\begin{tabular}[c]{@{}c@{}}BPCER@\\ APCER\\ 1\%\end{tabular}} & \multicolumn{1}{l|}{} & \multicolumn{1}{c|}{EER} \\ \cline{1-5} \cline{7-9} \cline{11-11} 
\multicolumn{1}{|c|}{-} & \multicolumn{1}{l|}{Baseline} & \multicolumn{1}{l|}{0.176} & \multicolumn{1}{l|}{0.24} & \multicolumn{1}{l|}{0.74} & \multicolumn{1}{l|}{} & \multicolumn{1}{l|}{0.1576} & \multicolumn{1}{l|}{0.4138} & \multicolumn{1}{l|}{1.000} & \multicolumn{1}{l|}{} & \multicolumn{1}{l|}{0.182} \\ \cline{1-5} \cline{7-9} \cline{11-11} 
\multicolumn{1}{|c|}{1} & \multicolumn{1}{l|}{MorphHRNet} & \multicolumn{1}{l|}{\textbf{0.042}} & \multicolumn{1}{l|}{0.112} & \multicolumn{1}{l|}{0.902} & \multicolumn{1}{l|}{} & \multicolumn{1}{l|}{0.0392} & \multicolumn{1}{l|}{0.1078} & \multicolumn{1}{l|}{0.5686} & \multicolumn{1}{l|}{} & \multicolumn{1}{l|}{0.098} \\ \cline{1-5} \cline{7-9} \cline{11-11} 
\multicolumn{1}{|c|}{2} & \multicolumn{1}{l|}{Xception} & \multicolumn{1}{l|}{0.108} & \multicolumn{1}{l|}{0.23} & \multicolumn{1}{l|}{0.494} & \multicolumn{1}{l|}{} & \multicolumn{1}{l|}{0.1176} & \multicolumn{1}{l|}{0.2157} & \multicolumn{1}{l|}{0.5392} & \multicolumn{1}{l|}{} & \multicolumn{1}{l|}{0.146} \\ \cline{1-5} \cline{7-9} \cline{11-11} 
\multicolumn{1}{|c|}{3} & \multicolumn{1}{l|}{Con-Text Net A} & \multicolumn{1}{l|}{0.31} & \multicolumn{1}{l|}{0.456} & \multicolumn{1}{l|}{0.892} & \multicolumn{1}{l|}{} & \multicolumn{1}{l|}{0.3186} & \multicolumn{1}{l|}{0.4853} & \multicolumn{1}{l|}{0.9314} & \multicolumn{1}{l|}{} & \multicolumn{1}{l|}{0.262} \\ \cline{1-5} \cline{7-9} \cline{11-11} 
\multicolumn{1}{|c|}{4} & \multicolumn{1}{l|}{Con-Text Net B} & \multicolumn{1}{l|}{0.438} & \multicolumn{1}{l|}{0.596} & \multicolumn{1}{l|}{0.81} & \multicolumn{1}{l|}{} & \multicolumn{1}{l|}{0.4412} & \multicolumn{1}{l|}{0.5539} & \multicolumn{1}{l|}{0.9461} & \multicolumn{1}{l|}{} & \multicolumn{1}{l|}{0.314} \\ \cline{1-5} \cline{7-9} \cline{11-11} 
\multicolumn{1}{|c|}{5} & \multicolumn{1}{l|}{E-CBAM@VCMI} & \multicolumn{1}{l|}{0.468} & \multicolumn{1}{l|}{0.868} & \multicolumn{1}{l|}{0.99} & \multicolumn{1}{l|}{} & \multicolumn{1}{l|}{0.3824} & \multicolumn{1}{l|}{0.4706} & \multicolumn{1}{l|}{0.6912} & \multicolumn{1}{l|}{} & \multicolumn{1}{l|}{0.306} \\ \cline{1-5} \cline{7-9} \cline{11-11} 
\multicolumn{1}{|c|}{6} & \multicolumn{1}{l|}{Anonymous} & \multicolumn{1}{l|}{1.000} & \multicolumn{1}{l|}{1.000} & \multicolumn{1}{l|}{1.000} & \multicolumn{1}{l|}{} & \multicolumn{1}{l|}{0.8627} & \multicolumn{1}{l|}{0.8775} & \multicolumn{1}{l|}{0.8971} & \multicolumn{1}{l|}{} & \multicolumn{1}{l|}{0.806} \\ \cline{1-5} \cline{7-9} \cline{11-11} 
\multicolumn{1}{|c|}{7} & \multicolumn{1}{l|}{VaMoS} & \multicolumn{1}{l|}{--} & \multicolumn{1}{l|}{--} & \multicolumn{1}{l|}{--} & \multicolumn{1}{l|}{} & \multicolumn{1}{l|}{} & \multicolumn{1}{l|}{} & \multicolumn{1}{l|}{} & \multicolumn{1}{l|}{} & \multicolumn{1}{l|}{--} \\ \cline{1-5} \cline{7-9} \cline{11-11} 
\end{tabular}}
\vspace{0.3mm}
\caption{The evaluation results on the dataset consisting of morphs created with the online tool Webmorph. Most systems had greater problems recognizing the morphs in comparison to the other landmark-based approaches.}
\label{webmorph}
\end{table*}

\begin{table*}[]
\centering
\small
\resizebox{0.85\textwidth}{!}{
\begin{tabular}{cllllllllll}
\multicolumn{1}{l}{} &  &  &  &  &  &  &  &  &  &  \\ \cline{2-11} 
\multicolumn{1}{l|}{} & \multicolumn{1}{c|}{\begin{tabular}[c]{@{}c@{}}Morphing\\ approach\end{tabular}} & \multicolumn{9}{c|}{MIPGAN-I} \\ \hline
\multicolumn{1}{|c|}{Rank} & \multicolumn{1}{c|}{Solution} & \multicolumn{1}{c|}{\textbf{\begin{tabular}[c]{@{}c@{}}APCER@\\ BPCER\\ 20\%\end{tabular}}} & \multicolumn{1}{c|}{\begin{tabular}[c]{@{}c@{}}APCER@\\ BPCER\\ 10\%\end{tabular}} & \multicolumn{1}{c|}{\begin{tabular}[c]{@{}c@{}}APCER@\\ BPCER\\ 1\%\end{tabular}} & \multicolumn{1}{c|}{} & \multicolumn{1}{c|}{\begin{tabular}[c]{@{}c@{}}BPCER@\\ APCER\\ 20\%\end{tabular}} & \multicolumn{1}{c|}{\begin{tabular}[c]{@{}c@{}}BPCER@\\ APCER\\ 10\%\end{tabular}} & \multicolumn{1}{c|}{\begin{tabular}[c]{@{}c@{}}BPCER@\\ APCER\\ 1\%\end{tabular}} & \multicolumn{1}{l|}{} & \multicolumn{1}{c|}{EER} \\ \cline{1-5} \cline{7-9} \cline{11-11} 
\multicolumn{1}{|c|}{-} & \multicolumn{1}{l|}{Baseline} & \multicolumn{1}{l|}{0.145} & \multicolumn{1}{l|}{0.222} & \multicolumn{1}{l|}{0.758} & \multicolumn{1}{l|}{} & \multicolumn{1}{l|}{0.1281} & \multicolumn{1}{l|}{0.33} & \multicolumn{1}{l|}{1.000} & \multicolumn{1}{l|}{} & \multicolumn{1}{l|}{0.167} \\ \cline{1-5} \cline{7-9} \cline{11-11} 
\multicolumn{1}{|c|}{1} & \multicolumn{1}{l|}{Con-Text Net A} & \multicolumn{1}{l|}{\textbf{0.081}} & \multicolumn{1}{l|}{0.141} & \multicolumn{1}{l|}{0.419} & \multicolumn{1}{l|}{} & \multicolumn{1}{l|}{0.0637} & \multicolumn{1}{l|}{0.1618} & \multicolumn{1}{l|}{0.5931} & \multicolumn{1}{l|}{} & \multicolumn{1}{l|}{0.123} \\ \cline{1-5} \cline{7-9} \cline{11-11} 
\multicolumn{1}{|c|}{2} & \multicolumn{1}{l|}{MorphHRNet} & \multicolumn{1}{l|}{0.13} & \multicolumn{1}{l|}{0.219} & \multicolumn{1}{l|}{0.898} & \multicolumn{1}{l|}{} & \multicolumn{1}{l|}{0.1127} & \multicolumn{1}{l|}{0.2402} & \multicolumn{1}{l|}{0.7598} & \multicolumn{1}{l|}{} & \multicolumn{1}{l|}{0.153} \\ \cline{1-5} \cline{7-9} \cline{11-11} 
\multicolumn{1}{|c|}{3} & \multicolumn{1}{l|}{Con-Text Net B} & \multicolumn{1}{l|}{0.398} & \multicolumn{1}{l|}{0.53} & \multicolumn{1}{l|}{0.715} & \multicolumn{1}{l|}{} & \multicolumn{1}{l|}{0.4559} & \multicolumn{1}{l|}{0.6029} & \multicolumn{1}{l|}{0.8775} & \multicolumn{1}{l|}{} & \multicolumn{1}{l|}{0.303} \\ \cline{1-5} \cline{7-9} \cline{11-11} 
\multicolumn{1}{|c|}{4} & \multicolumn{1}{l|}{Xception} & \multicolumn{1}{l|}{0.574} & \multicolumn{1}{l|}{0.804} & \multicolumn{1}{l|}{0.979} & \multicolumn{1}{l|}{} & \multicolumn{1}{l|}{0.4902} & \multicolumn{1}{l|}{0.5735} & \multicolumn{1}{l|}{0.8627} & \multicolumn{1}{l|}{} & \multicolumn{1}{l|}{0.369} \\ \cline{1-5} \cline{7-9} \cline{11-11} 
\multicolumn{1}{|c|}{5} & \multicolumn{1}{l|}{E-CBAM@VCMI} & \multicolumn{1}{l|}{0.602} & \multicolumn{1}{l|}{0.849} & \multicolumn{1}{l|}{0.999} & \multicolumn{1}{l|}{} & \multicolumn{1}{l|}{0.4069} & \multicolumn{1}{l|}{0.5343} & \multicolumn{1}{l|}{0.7892} & \multicolumn{1}{l|}{} & \multicolumn{1}{l|}{0.325} \\ \cline{1-5} \cline{7-9} \cline{11-11} 
\multicolumn{1}{|c|}{6} & \multicolumn{1}{l|}{Anonymous} & \multicolumn{1}{l|}{0.62} & \multicolumn{1}{l|}{0.756} & \multicolumn{1}{l|}{0.92} & \multicolumn{1}{l|}{} & \multicolumn{1}{l|}{0.4804} & \multicolumn{1}{l|}{0.4951} & \multicolumn{1}{l|}{0.5} & \multicolumn{1}{l|}{} & \multicolumn{1}{l|}{0.363} \\ \cline{1-5} \cline{7-9} \cline{11-11} 
\multicolumn{1}{|c|}{7} & \multicolumn{1}{l|}{VaMoS} & \multicolumn{1}{l|}{--} & \multicolumn{1}{l|}{--} & \multicolumn{1}{l|}{--} & \multicolumn{1}{l|}{} & \multicolumn{1}{l|}{0.902} & \multicolumn{1}{l|}{0.902} & \multicolumn{1}{l|}{0.9853} & \multicolumn{1}{l|}{} & \multicolumn{1}{l|}{--} \\ \cline{1-5} \cline{7-9} \cline{11-11} 
\end{tabular}}
\vspace{0.3mm}
\caption{The evaluation results on the GAN-based MIPGAN-I morph dataset. For the most systems, recognizing GAN-based morphs seems to be harder than recognizing landmark-based morphs.}
\label{mipgan1}
\end{table*}

\begin{table*}[]
\centering
\small
\resizebox{0.85\textwidth}{!}{
\begin{tabular}{cllllllllll}
\multicolumn{1}{l}{} &  &  &  &  &  &  &  &  &  &  \\ \cline{2-11} 
\multicolumn{1}{l|}{} & \multicolumn{1}{c|}{\begin{tabular}[c]{@{}c@{}}Morphing\\ approach\end{tabular}} & \multicolumn{9}{c|}{MIPGAN-II} \\ \hline
\multicolumn{1}{|c|}{Rank} & \multicolumn{1}{c|}{Solution} & \multicolumn{1}{c|}{\textbf{\begin{tabular}[c]{@{}c@{}}APCER@\\ BPCER\\ 20\%\end{tabular}}} & \multicolumn{1}{c|}{\begin{tabular}[c]{@{}c@{}}APCER@\\ BPCER\\ 10\%\end{tabular}} & \multicolumn{1}{c|}{\begin{tabular}[c]{@{}c@{}}APCER@\\ BPCER\\ 1\%\end{tabular}} & \multicolumn{1}{c|}{} & \multicolumn{1}{c|}{\begin{tabular}[c]{@{}c@{}}BPCER@\\ APCER\\ 20\%\end{tabular}} & \multicolumn{1}{c|}{\begin{tabular}[c]{@{}c@{}}BPCER@\\ APCER\\ 10\%\end{tabular}} & \multicolumn{1}{c|}{\begin{tabular}[c]{@{}c@{}}BPCER@\\ APCER\\ 1\%\end{tabular}} & \multicolumn{1}{l|}{} & \multicolumn{1}{c|}{EER} \\ \cline{1-5} \cline{7-9} \cline{11-11} 
\multicolumn{1}{|c|}{-} & \multicolumn{1}{l|}{Baseline} & \multicolumn{1}{l|}{0.2062} & \multicolumn{1}{l|}{0.3203} & \multicolumn{1}{l|}{0.8158} & \multicolumn{1}{l|}{} & \multicolumn{1}{l|}{0.2118} & \multicolumn{1}{l|}{0.3941} & \multicolumn{1}{l|}{1.000} & \multicolumn{1}{l|}{} & \multicolumn{1}{l|}{0.2062} \\ \cline{1-5} \cline{7-9} \cline{11-11} 
\multicolumn{1}{|c|}{1} & \multicolumn{1}{l|}{MorphHRNet} & \multicolumn{1}{l|}{\textbf{0.0611}} & \multicolumn{1}{l|}{0.1101} & \multicolumn{1}{l|}{0.8418} & \multicolumn{1}{l|}{} & \multicolumn{1}{l|}{0.0294} & \multicolumn{1}{l|}{0.1127} & \multicolumn{1}{l|}{0.6127} & \multicolumn{1}{l|}{} & \multicolumn{1}{l|}{0.1041} \\ \cline{1-5} \cline{7-9} \cline{11-11} 
\multicolumn{1}{|c|}{2} & \multicolumn{1}{l|}{Con-Text Net A} & \multicolumn{1}{l|}{0.0861} & \multicolumn{1}{l|}{0.1451} & \multicolumn{1}{l|}{0.4344} & \multicolumn{1}{l|}{} & \multicolumn{1}{l|}{0.0588} & \multicolumn{1}{l|}{0.1961} & \multicolumn{1}{l|}{0.5931} & \multicolumn{1}{l|}{} & \multicolumn{1}{l|}{0.1291} \\ \cline{1-5} \cline{7-9} \cline{11-11} 
\multicolumn{1}{|c|}{3} & \multicolumn{1}{l|}{E-CBAM@VCMI} & \multicolumn{1}{l|}{0.3734} & \multicolumn{1}{l|}{0.6466} & \multicolumn{1}{l|}{0.996} & \multicolumn{1}{l|}{} & \multicolumn{1}{l|}{0.3039} & \multicolumn{1}{l|}{0.3039} & \multicolumn{1}{l|}{0.5686} & \multicolumn{1}{l|}{} & \multicolumn{1}{l|}{0.2593} \\ \cline{1-5} \cline{7-9} \cline{11-11} 
\multicolumn{1}{|c|}{4} & \multicolumn{1}{l|}{Con-Text Net B} & \multicolumn{1}{l|}{0.3914} & \multicolumn{1}{l|}{0.5125} & \multicolumn{1}{l|}{0.6767} & \multicolumn{1}{l|}{} & \multicolumn{1}{l|}{0.4706} & \multicolumn{1}{l|}{0.6176} & \multicolumn{1}{l|}{0.9118} & \multicolumn{1}{l|}{} & \multicolumn{1}{l|}{0.2943} \\ \cline{1-5} \cline{7-9} \cline{11-11} 
\multicolumn{1}{|c|}{5} & \multicolumn{1}{l|}{Anonymous} & \multicolumn{1}{l|}{0.7548} & \multicolumn{1}{l|}{0.8539} & \multicolumn{1}{l|}{0.975} & \multicolumn{1}{l|}{} & \multicolumn{1}{l|}{0.4951} & \multicolumn{1}{l|}{0.5000} & \multicolumn{1}{l|}{0.5637} & \multicolumn{1}{l|}{} & \multicolumn{1}{l|}{0.4334} \\ \cline{1-5} \cline{7-9} \cline{11-11} 
\multicolumn{1}{|c|}{6} & \multicolumn{1}{l|}{Xception} & \multicolumn{1}{l|}{0.7708} & \multicolumn{1}{l|}{0.9249} & \multicolumn{1}{l|}{0.995} & \multicolumn{1}{l|}{} & \multicolumn{1}{l|}{0.5686} & \multicolumn{1}{l|}{0.6765} & \multicolumn{1}{l|}{0.902} & \multicolumn{1}{l|}{} & \multicolumn{1}{l|}{0.4454} \\ \cline{1-5} \cline{7-9} \cline{11-11} 
\multicolumn{1}{|c|}{7} & \multicolumn{1}{l|}{VaMoS} & \multicolumn{1}{l|}{--} & \multicolumn{1}{l|}{--} & \multicolumn{1}{l|}{--} & \multicolumn{1}{l|}{} & \multicolumn{1}{l|}{0.902} & \multicolumn{1}{l|}{0.902} & \multicolumn{1}{l|}{0.9314} & \multicolumn{1}{l|}{} & \multicolumn{1}{l|}{--} \\ \cline{1-5} \cline{7-9} \cline{11-11} 
\end{tabular}}
\vspace{0.3mm}
\caption{The evaluation results on the GAN-based MIPGAN-II morph dataset. Similar to the results on the MIPGAN-I dataset, recognizing GAN-based morphs seems to be harder than recognizing landmark-based morphs for the most face morphing detector systems.}
\label{mipgan2}
\end{table*}

\vspace{-3mm}
\section{Results and Analysis} 
\vspace{-2.5mm}
This section presents the evaluation results of the submitted solutions and the baseline approach on the five different morphing approach datasets presented in Section \ref{morphing} in terms of the BPCER at a fixed APCER and the performance in terms of APCER at a fixed BPCER. We furthermore also report the detection Equal Error Rate (EER). We first present and discuss the results on the Landmark-based morphing approach datasets (OpenCV, FaceMorpher, and Webmorph) and then on the GAN-based morphing approach datasets (MIPGAN-I and MIPGAN-II). The final ranking of the submitted solutions is shown in Table \ref{ranking}. None of the submitted solutions consistently outperformed all other solutions on all benchmarks. Considering the average ranking on all benchmarks, MorphHRNet achieved the top-1 rank.

\vspace{-1.8mm}
\subsection{Landmark-based Morphing Approaches}
\vspace{-1.6mm}
The results on the landmark-based morphing approaches of the submitted solutions are shown in Table \ref{opencv} for the morphs based on the OpenCV approach, in Table \ref{facemorpher} for the FaceMorpher-based morphs, and in Table \ref{webmorph} for the morphs created using the Webmorph online tool.

The \textbf{MorphHRNet} and the \textbf{Xception} model outperform the baseline on all three landmark-based morph datasets in terms of APCER@BPCER=20\%. On the OpenCV and the Webmorph benchmarks (Table \ref{opencv}, \ref{webmorph}) they are the top-2 performing submissions. On the Webmorph dataset, the MorphHRNet significantly outperformed the other submitted approaches. Both \textbf{Con-Text Net} models, \textbf{A} and \textbf{B} are top-performing solutions on the FaceMorpher dataset in terms of APCER@BPCER=20\% (Table \ref{facemorpher}).  On the other landmark-based morph datasets, Webmorph (Table \ref{webmorph}) and OpenCV (Table \ref{opencv}), they achieved lower performance than MorphHRNet, Xception, and the baseline. The \textbf{E-CBAM@VCMI model} is ranked behind MorphHRNet, Xception, Con-Text Net A and B, and achieved lower performance than the baseline on the landmark-based morphing datasets as shown in Tables \ref{opencv}, \ref{facemorpher}, \ref{webmorph}. The \textbf{Anonymous} model, based on hand-crafted features, shows interesting behavior, as it shows only some ability to distinguish between morphs and bona fide on the OpenCV dataset. This behavior has been predicted by the team, as their approach has been specialized based on the provided synthetic morphs, which were created using the OpenCV morphing approach. 
The evaluation of the \textbf{VaMoS} approach was impossible, especially when fixing BPCER.
This is due to the fact that almost for all bona fide samples, and most of the attacks, the solution resulted in the same exact detection score, which makes setting up a threshold at a specific BPCER impossible. In such situations, the results in the relative tables are marked with "-".
On the test data used by the submitting team, the organizers were able to reproduce the desired results. The team's assumption is strong over-fitting of the model, but without deeper insights into the specific details, the organizing team cannot make a more specific statement.
\begin{table*}[]
\small
\centering
\begin{tabular}{lccccccc}
\hline
 & \multicolumn{7}{c}{Ranks} \\ \hline
Solution & OpenCV & FaceMorpher & Webmorph & MIPGAN-I & MIPGAN-II & Avg & Final rank \\ \hline
\multicolumn{1}{l|}{MorphHRNet} & 1 & 4 & 1 & 2 & \multicolumn{1}{c|}{1} & \multicolumn{1}{c|}{1.8} & 1 \\
\multicolumn{1}{l|}{Con-Text Net A} & 3 & 3 & 3 & 1 & \multicolumn{1}{c|}{2} & \multicolumn{1}{c|}{2.4} & 2 \\
\multicolumn{1}{l|}{Xception} & 2 & 2 & 2 & 4 & \multicolumn{1}{c|}{6} & \multicolumn{1}{c|}{3.2} & 3 \\
\multicolumn{1}{l|}{Con-Text Net B} & 4 & 1 & 4 & 3 & \multicolumn{1}{c|}{4} & \multicolumn{1}{c|}{3.2} & 3 \\
\multicolumn{1}{l|}{E-CBAM@VCMI} & 5 & 5 & 5 & 5 & \multicolumn{1}{c|}{3} & \multicolumn{1}{c|}{4.6} & 4 \\
\multicolumn{1}{l|}{Anonymous} & 6 & 6 & 6 & 6 & \multicolumn{1}{c|}{5} & \multicolumn{1}{c|}{5.8} & 5 \\
\multicolumn{1}{l|}{VaMoS} & 7 & 7 & 7 & 7 & \multicolumn{1}{c|}{7} & \multicolumn{1}{c|}{7} & 6 \\
\hline
\end{tabular}
\vspace{0.3mm}
\caption{The final ranking of the submitted solutions based on the average rank of the performance on the five morphing datasets.}
\label{ranking}
\vspace{-2.5mm}
\end{table*}

\vspace{-1.8mm}
\subsection{GAN-based Morphing Approaches}
\vspace{-1.5mm}
The results on the GAN-based morphing approaches of the submitted solutions are shown in Table \ref{mipgan1} for the morphs based on the MIPGAN-I morphing approach and in Table \ref{mipgan2} for the MIPGAN-II-based morphs.

MorphHRNet achieved the best performance on MIPGAN-II and ranked second on MIPGAN-I, as shown in Table \ref{mipgan1}, \ref{mipgan2}. Xception, which achieved very competitive results on landmark-based morphing datasets, achieved relatively very low performance on GAN-based morphing datasets. Similar behavior can be seen for the \textbf{Con-Text Net} models. While the Con-Text Net model trained on less epochs, \textbf{A}, outperforms the baseline on both datasets, the model trained for more epochs (\textbf{B}) struggles to reach the baseline performance. The performance of the \textbf{E-CBAM@VCMI} approach, as well as the \textbf{Anonymous} hand-crafted approach, does not reach the baseline performance, although the E-CBAM@VMCI model outperforms the Xception and the Con-Text Net B model on the MIPGAN-II dataset. The \textbf{VaMoS} approach did again not allow an evaluation at APCER@BPCER=20\% or any other operational points.

\vspace{-1.5mm}
\subsection{Comparison and Final Ranking} 
\vspace{-1.5mm}
In this subsection, we briefly investigate the performance differences of the submitted solutions on landmark-based and deep learning-based morphing approaches and also investigate the final ranking based on the average rank achieved on all five morphing benchmarks. The final ranking is presented in Table \ref{ranking}.

For the \textbf{MorphHRNet} approach, we can observe that the model performs well independently of the used morphing approach, both, landmark-based and GAN-based face morphs are relatively well detected by the model. The \textbf{Con-Text Net A} model especially performs well on GAN-based face morphs and lacks some performance on landmark-based morphs. The \textbf{Xception} model, which showed good performance on the landmark-based approach, did not classify the GAN-based morphs well, as can be observed in Tables \ref{mipgan1} and \ref{mipgan2}. Similar behavior can be observed for the \textbf{Con-Text Net B} model. For the \textbf{E-CBAM@VCMI} model, no specific change in performance was observable regarding being used for either type of morphing approaches, landmark, or GAN-based. No particular behavior can be observed with regard to the performance of \textbf{Anonymous} and \textbf{VaMoS} besides their lack of power to distinguish between morphs and bona fide face images.

In the final ranking, the \textbf{MorphHRNet} won the competition with an average rank of 1.8, the second place was achieved by the \textbf{Con-Text Net A} model (average rank 2.4), and the third place is shared between the \textbf{Xception} model and the \textbf{Con-Text Net B} model (both average rank 3.2), as detailed in Table \ref{ranking}.

\vspace{-1.5mm}
\section{Conclusion} 
\vspace{-1.5mm}
In this paper, we summarized the results and observations of the SYN-MAD 2022: Competition on Face Morphing Attack Detection Based on Privacy-aware Synthetic Training Data. In total, 12 teams registered for participation, and six of them submitted seven valid submissions to tackle the problem of face morphing detection while considering the privacy and legal issues related to real development data. The evaluation focused on a wide range of different morphing approaches, including landmark- and GAN-based approaches, and various creative solutions have been evaluated, leading to enhanced performances in comparison to the considered baseline. 

\vspace{-1.5mm}
\subsection*{Acknowledgments}
\vspace{-1.5mm}
This research work has been funded by the German Federal Ministry of Education and Research and the Hessen State Ministry for Higher Education, Research and the Arts within their joint support of the National Research Center for Applied Cybersecurity ATHENE.

{\small
\bibliographystyle{ieee}
\bibliography{egbib}

\begin{thebibliography}{10}\itemsep=-1pt

\bibitem{DBLP:conf/icb/BoutrosDFKK21}
F.~Boutros, N.~Damer, M.~Fang, F.~Kirchbuchner, and A.~Kuijper.
\newblock Mixfacenets: Extremely efficient face recognition networks.
\newblock In {\em {IJCB}}, pages 1--8. {IEEE}, 2021.

\bibitem{DBLP:journals/corr/abs-2109-09416}
F.~Boutros, N.~Damer, F.~Kirchbuchner, and A.~Kuijper.
\newblock Elasticface: Elastic margin loss for deep face recognition.
\newblock In {\em Proceedings of the IEEE/CVF Conference on Computer Vision and
  Pattern Recognition (CVPR) Workshops}, pages 1578--1587, June 2022.

\bibitem{DBLP:journals/corr/abs-2112-06592}
F.~Boutros, M.~Fang, M.~Klemt, B.~Fu, and N.~Damer.
\newblock {CR-FIQA:} face image quality assessment by learning sample relative
  classifiability.
\newblock {\em CoRR}, abs/2112.06592, 2021.

\bibitem{carabemethods}
L.~C{\'a}rabe and E.~Cermeno.
\newblock Methods to prevent registration using fake face images.

\bibitem{DBLP:journals/access/CarabeC21}
L.~C{\'{a}}rabe and E.~Cerme{\~{n}}o.
\newblock Stegano-morphing: Concealing attacks on face identification
  algorithms.
\newblock {\em {IEEE} Access}, 9:100851--100867, 2021.

\bibitem{DBLP:conf/icb/ChingovskaYLYLKGDKNKPGKBRKGWZAKGTPSRGFPPSRAM13}
I.~Chingovska, J.~Yang, Z.~Lei, D.~Yi, S.~Z. Li, O.~K{\"{a}}hm, C.~Glaser,
  N.~Damer, A.~Kuijper, A.~Nouak, J.~Komulainen, T.~F. Pereira, S.~Gupta,
  S.~Khandelwal, S.~Bansal, A.~Rai, T.~Krishna, D.~Goyal, M.~Waris, H.~Zhang,
  I.~Ahmad, S.~Kiranyaz, M.~Gabbouj, R.~Tronci, M.~Pili, N.~Sirena, F.~Roli,
  J.~Galbally, J.~Fi{\'{e}}rrez, A.~da~Silva~Pinto, H.~Pedrini, W.~S. Schwartz,
  A.~Rocha, A.~Anjos, and S.~Marcel.
\newblock The 2nd competition on counter measures to 2d face spoofing attacks.
\newblock In {\em {ICB}}, pages 1--6. {IEEE}, 2013.

\bibitem{DBLP:conf/cvpr/Chollet17}
F.~Chollet.
\newblock Xception: Deep learning with depthwise separable convolutions.
\newblock In {\em {CVPR}}, pages 1800--1807. {IEEE} Computer Society, 2017.

\bibitem{DBLP:conf/dagm/DamerBWBTBK18}
N.~Damer, V.~Boller, Y.~Wainakh, F.~Boutros, P.~Terh{\"{o}}rst, A.~Braun, and
  A.~Kuijper.
\newblock Detecting face morphing attacks by analyzing the directed distances
  of facial landmarks shifts.
\newblock In {\em {GCPR}}, volume 11269 of {\em Lecture Notes in Computer
  Science}, pages 518--534. Springer, 2018.

\bibitem{DBLP:conf/btas/DamerBSKK19}
N.~Damer, F.~Boutros, A.~M. Saladie, F.~Kirchbuchner, and A.~Kuijper.
\newblock Realistic dreams: Cascaded enhancement of gan-generated images with
  an example in face morphing attacks.
\newblock In {\em {BTAS}}, pages 1--10. {IEEE}, 2019.

\bibitem{DBLP:conf/btas/DamerGZKK19}
N.~Damer, J.~H. Grebe, S.~Zienert, F.~Kirchbuchner, and A.~Kuijper.
\newblock On the generalization of detecting face morphing attacks as
  anomalies: Novelty vs. outlier detection.
\newblock In {\em {BTAS}}, pages 1--5. {IEEE}, 2019.

\bibitem{DBLP:journals/corr/abs-2203-06691}
N.~Damer, C.~A.~F. L\'opez, M.~Fang, N.~Spiller, M.~V. Pham, and F.~Boutros.
\newblock Privacy-friendly synthetic data for the development of face morphing
  attack detectors.
\newblock In {\em Proceedings of the IEEE/CVF Conference on Computer Vision and
  Pattern Recognition (CVPR) Workshops}, pages 1606--1617, June 2022.

\bibitem{DBLP:conf/isvc/DamerRSVBFKRK21}
N.~Damer, K.~B. Raja, M.~S{\"{u}}{\ss}milch, S.~Venkatesh, F.~Boutros, M.~Fang,
  F.~Kirchbuchner, R.~Ramachandra, and A.~Kuijper.
\newblock Regenmorph: Visibly realistic {GAN} generated face morphing attacks
  by attack re-generation.
\newblock In {\em {ISVC} {(1)}}, volume 13017 of {\em Lecture Notes in Computer
  Science}, pages 251--264. Springer, 2021.

\bibitem{DBLP:conf/btas/DamerS0K18}
N.~Damer, A.~M. Saladie, A.~Braun, and A.~Kuijper.
\newblock Morgan: Recognition vulnerability and attack detectability of face
  morphing attacks created by generative adversarial network.
\newblock In {\em {BTAS}}, pages 1--10. {IEEE}, 2018.

\bibitem{DBLP:conf/icb/DamerSZWTKK19}
N.~Damer, A.~M. Saladie, S.~Zienert, Y.~Wainakh, P.~Terh{\"{o}}rst,
  F.~Kirchbuchner, and A.~Kuijper.
\newblock To detect or not to detect: The right faces to morph.
\newblock In {\em {ICB}}, pages 1--8. {IEEE}, 2019.

\bibitem{DBLP:conf/isvc/DamerSFBKK21}
N.~Damer, N.~Spiller, M.~Fang, F.~Boutros, F.~Kirchbuchner, and A.~Kuijper.
\newblock {PW-MAD:} pixel-wise supervision for generalized face morphing attack
  detection.
\newblock In {\em {ISVC} {(1)}}, volume 13017 of {\em Lecture Notes in Computer
  Science}, pages 291--304. Springer, 2021.

\bibitem{DBLP:conf/iciap/DebiasiDSRSBKU19}
L.~Debiasi, N.~Damer, A.~M. Saladie, C.~Rathgeb, U.~Scherhag, C.~Busch,
  F.~Kirchbuchner, and A.~Uhl.
\newblock On the detection of gan-based face morphs using established morph
  detectors.
\newblock In {\em {ICIAP} {(2)}}, volume 11752 of {\em Lecture Notes in
  Computer Science}, pages 345--356. Springer, 2019.

\bibitem{DeBruine2021}
L.~DeBruine and B.~Jones.
\newblock {Face Research Lab London Set}.
\newblock 4 2021.

\bibitem{DBLP:conf/cvpr/DengGVKZ20}
J.~Deng, J.~Guo, E.~Ververas, I.~Kotsia, and S.~Zafeiriou.
\newblock Retinaface: Single-shot multi-level face localisation in the wild.
\newblock In {\em {CVPR}}, pages 5202--5211. Computer Vision Foundation /
  {IEEE}, 2020.

\bibitem{DBLP:conf/cvpr/DengGXZ19}
J.~Deng, J.~Guo, N.~Xue, and S.~Zafeiriou.
\newblock Arcface: Additive angular margin loss for deep face recognition.
\newblock In {\em {CVPR}}, pages 4690--4699. Computer Vision Foundation /
  {IEEE}, 2019.

\bibitem{DBLP:journals/corr/ElfwingUD17}
S.~Elfwing, E.~Uchibe, and K.~Doya.
\newblock Sigmoid-weighted linear units for neural network function
  approximation in reinforcement learning.
\newblock {\em CoRR}, abs/1702.03118, 2017.

\bibitem{DBLP:conf/fgr/FangBKD21}
M.~Fang, F.~Boutros, A.~Kuijper, and N.~Damer.
\newblock Partial attack supervision and regional weighted inference for masked
  face presentation attack detection.
\newblock In {\em {FG}}, pages 1--8. {IEEE}, 2021.

\bibitem{DBLP:conf/icb/FerraraFM14}
M.~Ferrara, A.~Franco, and D.~Maltoni.
\newblock The magic passport.
\newblock In {\em {IJCB}}, pages 1--7. {IEEE}, 2014.

\bibitem{DBLP:journals/tifs/FerraraFM18}
M.~Ferrara, A.~Franco, and D.~Maltoni.
\newblock Face demorphing.
\newblock {\em {IEEE} Trans. Inf. Forensics Secur.}, 13(4):1008--1017, 2018.

\bibitem{DBLP:conf/cvpr/HeZRS16}
K.~He, X.~Zhang, S.~Ren, and J.~Sun.
\newblock Deep residual learning for image recognition.
\newblock In {\em {CVPR}}, pages 770--778. {IEEE} Computer Society, 2016.

\bibitem{LFWTech}
G.~B. Huang, M.~Ramesh, T.~Berg, and E.~Learned-Miller.
\newblock Labeled faces in the wild: A database for studying face recognition
  in unconstrained environments.
\newblock Technical Report 07-49, University of Massachusetts, Amherst, October
  2007.

\bibitem{DBLP:conf/cvpr/HuangWT0SLLH20}
Y.~Huang, Y.~Wang, Y.~Tai, X.~Liu, P.~Shen, S.~Li, J.~Li, and F.~Huang.
\newblock Curricularface: Adaptive curriculum learning loss for deep face
  recognition.
\newblock In {\em 2020 {IEEE/CVF} Conference on Computer Vision and Pattern
  Recognition, {CVPR} 2020, Seattle, WA, USA, June 13-19, 2020}, pages
  5900--5909. Computer Vision Foundation / {IEEE}, 2020.

\bibitem{ISO301073}
{International Organization for Standardization}.
\newblock {ISO/IEC DIS 30107-3:2016: Information Technology – Biometric
  presentation attack detection – P. 3: Testing and reporting}, 2017.

\bibitem{ISOIEC29794-1}
{ISO/IEC JTC1 SC37 Biometrics}.
\newblock {ISO/IEC 29794-1:2016 Information technology - Biometric sample
  quality - Part 1: Framework}.
\newblock International Organization for Standardization, 2016.

\bibitem{batagelj2022morphing}
M.~Ivanovska, A.~Kronov{\v{s}}ek, P.~Peer, V.~{\v{S}}truc, and B.~Batagelj.
\newblock Face morphing attack detection using privacy-aware training data.
\newblock In {\em Proceedings of the 31st International Electrotechnical and
  Computer Science Conference (under review)}, 2022.

\bibitem{DBLP:conf/nips/KarrasAHLLA20}
T.~Karras, M.~Aittala, J.~Hellsten, S.~Laine, J.~Lehtinen, and T.~Aila.
\newblock Training generative adversarial networks with limited data.
\newblock In {\em NeurIPS}, 2020.

\bibitem{DBLP:conf/cvpr/KarrasLA19}
T.~Karras, S.~Laine, and T.~Aila.
\newblock A style-based generator architecture for generative adversarial
  networks.
\newblock In {\em {CVPR}}, pages 4401--4410. Computer Vision Foundation /
  {IEEE}, 2019.

\bibitem{DBLP:conf/cvpr/KarrasLAHLA20}
T.~Karras, S.~Laine, M.~Aittala, J.~Hellsten, J.~Lehtinen, and T.~Aila.
\newblock Analyzing and improving the image quality of stylegan.
\newblock In {\em 2020 {IEEE/CVF} Conference on Computer Vision and Pattern
  Recognition, {CVPR} 2020, Seattle, WA, USA, June 13-19, 2020}, pages
  8107--8116. Computer Vision Foundation / {IEEE}, 2020.

\bibitem{DBLP:conf/cvpr/KazemiS14}
V.~Kazemi and J.~Sullivan.
\newblock One millisecond face alignment with an ensemble of regression trees.
\newblock In {\em 2014 {IEEE} Conference on Computer Vision and Pattern
  Recognition, {CVPR} 2014, Columbus, OH, USA, June 23-28, 2014}, pages
  1867--1874. {IEEE} Computer Society, 2014.

\bibitem{DBLP:journals/jmlr/King09}
D.~E. King.
\newblock Dlib-ml: {A} machine learning toolkit.
\newblock {\em J. Mach. Learn. Res.}, 10:1755--1758, 2009.

\bibitem{DBLP:journals/corr/KingmaB14}
D.~P. Kingma and J.~Ba.
\newblock Adam: {A} method for stochastic optimization.
\newblock In {\em {ICLR} (Poster)}, 2015.

\bibitem{openCVmorph}
S.~Mallick.
\newblock Face morph using opencv — c++ / python.
\newblock {\em LearnOpenCV}, 1(1), 2016.

\bibitem{DBLP:journals/tifs/MedenRTDKSRPS21}
B.~Meden, P.~Rot, P.~Terh{\"{o}}rst, N.~Damer, A.~Kuijper, W.~J. Scheirer,
  A.~Ross, P.~Peer, and V.~Struc.
\newblock Privacy-enhancing face biometrics: {A} comprehensive survey.
\newblock {\em {IEEE} Trans. Inf. Forensics Secur.}, 16:4147--4183, 2021.

\bibitem{DBLP:conf/iwdw/Neubert17}
T.~Neubert.
\newblock Face morphing detection: An approach based on image degradation
  analysis.
\newblock In {\em {IWDW}}, volume 10431 of {\em Lecture Notes in Computer
  Science}, pages 93--106. Springer, 2017.

\bibitem{NIST_Morph}
M.~Ngan, P.~Grother, K.~Hanaoka, and J.~Kuo.
\newblock {Face Recognition Vendor Test (FRVT) Part 4: MORPH - Performance of
  Automated Face Morph Detection}.
\newblock {\em {National Institute of Standards and Technology (NIST)}}, 2021.

\bibitem{DBLP:conf/icb/PurnapatraSBDYM21}
S.~Purnapatra, N.~Smalt, K.~Bahmani, P.~Das, D.~Yambay, A.~Mohammadi,
  A.~George, T.~Bourlai, S.~Marcel, S.~Schuckers, M.~Fang, N.~Damer,
  F.~Boutros, A.~Kuijper, A.~Kantarci, B.~Demir, Z.~Yildiz, Z.~Ghafoory,
  H.~Dertli, H.~K. Ekenel, S.~Vu, V.~Christophides, D.~Liang, G.~Zhang, Z.~Hao,
  J.~Liu, Y.~Jin, S.~Liu, S.~Huang, S.~Kuei, J.~M. Singh, and R.~Ramachandra.
\newblock Face liveness detection competition (livdet-face) - 2021.
\newblock In {\em {IJCB}}, pages 1--10. {IEEE}, 2021.

\bibitem{DBLP:conf/iccv/QiuYG00T21}
H.~Qiu, B.~Yu, D.~Gong, Z.~Li, W.~Liu, and D.~Tao.
\newblock Synface: Face recognition with synthetic data.
\newblock In {\em {ICCV}}, pages 10860--10870. {IEEE}, 2021.

\bibitem{DBLP:conf/btas/RaghavendraRB16}
R.~Raghavendra, K.~B. Raja, and C.~Busch.
\newblock Detecting morphed face images.
\newblock In {\em {BTAS}}, pages 1--7. {IEEE}, 2016.

\bibitem{DBLP:conf/icb/RaghavendraRVB17}
R.~Raghavendra, K.~B. Raja, S.~Venkatesh, and C.~Busch.
\newblock Face morphing versus face averaging: Vulnerability and detection.
\newblock In {\em {IJCB}}, pages 555--563. {IEEE}, 2017.

\bibitem{DBLP:conf/cvpr/RaghavendraRVB17a}
R.~Raghavendra, K.~B. Raja, S.~Venkatesh, and C.~Busch.
\newblock Transferable deep-cnn features for detecting digital and
  print-scanned morphed face images.
\newblock In {\em {CVPR} Workshops}, pages 1822--1830. {IEEE} Computer Society,
  2017.

\bibitem{DBLP:journals/corr/abs-2102-00813}
I.~D. Raji and G.~Fried.
\newblock About face: {A} survey of facial recognition evaluation.
\newblock {\em CoRR}, abs/2102.00813, 2021.

\bibitem{DBLP:conf/iccv/RosslerCVRTN19}
A.~R{\"{o}}ssler, D.~Cozzolino, L.~Verdoliva, C.~Riess, J.~Thies, and
  M.~Nie{\ss}ner.
\newblock Faceforensics++: Learning to detect manipulated facial images.
\newblock In {\em {ICCV}}, pages 1--11. {IEEE}, 2019.

\bibitem{DBLP:journals/corr/abs-2012-05344}
E.~Sarkar, P.~Korshunov, L.~Colbois, and S.~Marcel.
\newblock Vulnerability analysis of face morphing attacks from landmarks and
  generative adversarial networks.
\newblock {\em CoRR}, abs/2012.05344, 2020.

\bibitem{DBLP:conf/biosig/ScherhagNRGVSSM17}
U.~Scherhag, A.~Nautsch, C.~Rathgeb, M.~Gomez{-}Barrero, R.~N.~J. Veldhuis,
  L.~J. Spreeuwers, M.~Schils, D.~Maltoni, P.~Grother, S.~Marcel,
  R.~Breithaupt, R.~Raghavendra, and C.~Busch.
\newblock Biometric systems under morphing attacks: Assessment of morphing
  techniques and vulnerability reporting.
\newblock In {\em International Conference of the Biometrics Special Interest
  Group, {BIOSIG} 2017, Darmstadt, Germany, September 20-22, 2017}, volume
  {P-270} of {\em {LNI}}, pages 149--159. {GI} / {IEEE}, 2017.

\bibitem{DBLP:journals/tifs/ScherhagRMB20}
U.~Scherhag, C.~Rathgeb, J.~Merkle, and C.~Busch.
\newblock Deep face representations for differential morphing attack detection.
\newblock {\em {IEEE} Trans. Inf. Forensics Secur.}, 15:3625--3639, 2020.

\bibitem{DBLP:conf/cvpr/SchroffKP15}
F.~Schroff, D.~Kalenichenko, and J.~Philbin.
\newblock Facenet: {A} unified embedding for face recognition and clustering.
\newblock In {\em {CVPR}}, pages 815--823. {IEEE} Computer Society, 2015.

\bibitem{DBLP:conf/cvpr/0009XLW19}
K.~Sun, B.~Xiao, D.~Liu, and J.~Wang.
\newblock Deep high-resolution representation learning for human pose
  estimation.
\newblock In {\em {CVPR}}, pages 5693--5703. Computer Vision Foundation /
  {IEEE}, 2019.

\bibitem{DBLP:conf/iwbf/VenkateshZRRDB20}
S.~Venkatesh, H.~Zhang, R.~Ramachandra, K.~B. Raja, N.~Damer, and C.~Busch.
\newblock Can {GAN} generated morphs threaten face recognition systems equally
  as landmark based morphs? - vulnerability and detection.
\newblock In {\em {IWBF}}, pages 1--6. {IEEE}, 2020.

\bibitem{gdpr}
P.~Voigt and A.~v.~d. Bussche.
\newblock {\em The EU General Data Protection Regulation (GDPR): A Practical
  Guide}.
\newblock Springer, 1st edition, 2017.

\bibitem{DBLP:journals/pami/00010CJDZ0MTW0X21}
J.~Wang, K.~Sun, T.~Cheng, B.~Jiang, C.~Deng, Y.~Zhao, D.~Liu, Y.~Mu, M.~Tan,
  X.~Wang, W.~Liu, and B.~Xiao.
\newblock Deep high-resolution representation learning for visual recognition.
\newblock {\em {IEEE} Trans. Pattern Anal. Mach. Intell.}, 43(10):3349--3364,
  2021.

\bibitem{DBLP:conf/eccv/WooPLK18}
S.~Woo, J.~Park, J.~Lee, and I.~S. Kweon.
\newblock {CBAM:} convolutional block attention module.
\newblock In {\em {ECCV} {(7)}}, volume 11211 of {\em Lecture Notes in Computer
  Science}, pages 3--19. Springer, 2018.

\bibitem{DBLP:journals/tbbis/ZhangVRRDB21}
H.~Zhang, S.~Venkatesh, R.~Ramachandra, K.~B. Raja, N.~Damer, and C.~Busch.
\newblock {MIPGAN} - generating strong and high quality morphing attacks using
  identity prior driven {GAN}.
\newblock {\em {IEEE} Trans. Biom. Behav. Identity Sci.}, 3(3):365--383, 2021.

\end{thebibliography}
}

\end{document}